\documentclass[times,twocolumn,final]{elsarticle}
\usepackage{medima}
\usepackage{framed,multirow}

\usepackage{amssymb}
\usepackage{latexsym}

\usepackage{graphicx}
\usepackage{glossaries}
\usepackage{algorithm}
\usepackage{algorithmic}
\usepackage{amsthm}

\usepackage{amsmath}  
\usepackage{bbm}
\usepackage{multirow}
\usepackage{booktabs}
\usepackage{float}
\usepackage{xcolor}
\usepackage{caption}
\usepackage{threeparttable, tablefootnote}
\usepackage[font=small]{subcaption}
\usepackage{soul}
\usepackage{lineno}
\usepackage{tabularx}

\usepackage{fontawesome}
\floatname{algorithm}{Algorithm}  

\usepackage{url}
\usepackage{xcolor}

\usepackage[colorlinks]{hyperref}

\definecolor{newcolor}{rgb}{.8,.349,.1}
\definecolor{navyblue}{HTML}{1155cc}

\newcommand{\argmin}[1]{\underset{#1}{\operatorname{arg}\,\operatorname{min}}\;}

\usepackage[normalem]{ulem}

\journal{Medical Image Analysis}

\begin{document}

\verso{T.L. Bobrow \textit{et~al.}}

\begin{frontmatter}

\title{Colonoscopy 3D video dataset with paired depth from 2D-3D registration}

\author[1]{Taylor L. \snm{Bobrow}}
\author[1]{Mayank \snm{Golhar}}
\author[1]{Rohan \snm{Vijayan}}
\author[2]{Venkata S. \snm{Akshintala}}
\author[3]{Juan R. \snm{Garcia}}
\author[1]{Nicholas J. \snm{Durr}\corref{cor1}} \ead{ndurr@jhu.edu}

\address[1]{Department of Biomedical Engineering, Johns Hopkins University, Baltimore, MD 21218, USA}
\address[2]{Division of Gastroenterology and Hepatology, Johns Hopkins Medicine, Baltimore, MD 21287, USA}
\address[3]{Department of Art as Applied to Medicine, Johns Hopkins School of Medicine, Baltimore, MD 21287, USA}

\cortext[cor1]{Corresponding author:}

\received{Nov 2022}
\accepted{TBD}
\availableonline{TBD}


\begin{abstract}
\centering\begin{minipage}{\dimexpr\paperwidth-7cm}
Screening colonoscopy is an important clinical application for several 3D computer vision techniques, including depth estimation, surface reconstruction, and missing region detection. However, the development, evaluation, and comparison of these techniques in real colonoscopy videos remain largely qualitative due to the difficulty of acquiring ground truth data. In this work, we present a \underline{C}olonoscopy \underline{3}D \underline{V}ideo \underline{D}ataset (C3VD) acquired with a high definition clinical colonoscope and high-fidelity colon models for benchmarking computer vision methods in colonoscopy. We introduce a novel multimodal 2D-3D registration technique to register optical video sequences with ground truth rendered views of a known 3D model. The different modalities are registered by transforming optical images to depth maps with a Generative Adversarial Network and aligning edge features with an evolutionary optimizer. This registration method achieves an average translation error of 0.321 millimeters and an average rotation error of 0.159 degrees in simulation experiments where error-free ground truth is available. The method also leverages video information, improving registration accuracy by 55.6\% for translation and 60.4\% for rotation compared to single frame registration. 22 short video sequences were registered to generate 10,015 total frames with paired ground truth depth, surface normals, optical flow, occlusion, six degree-of-freedom pose, coverage maps, and 3D models. The dataset also includes screening videos acquired by a gastroenterologist with paired ground truth pose and 3D surface models. The dataset and registration source code are available at \url{durr.jhu.edu/C3VD}.
\end{minipage}
\end{abstract}
\end{frontmatter}

\section{Introduction}

Colorectal cancer (CRC) is the second most lethal form of cancer in the United States \citep{siegel2021}. At least 80\% of CRCs are believed to develop from premalignant adenomas \citep{cunningham2010}. Screening colonoscopy remains the gold-standard for detecting and removing precancerous lesions, effectively reducing the risk of developing CRC \citep{rex2015}. Still, an estimated 22\% of precancerous lesions go undetected during screening procedures \citep{rijn2006}. These missed lesions are thought to be a primary contributor to interval CRC --- the development of CRC within 5 years of a negative screening colonoscopy --- which represents 6\% of CRC cases \citep{samadder2014}. 

Colonoscopy remains an active application for computer vision researchers working to reduce lesion miss rates and improve clinical outcomes \citep{ali2020,fu2021,chadebecq2023}. Recent works have employed Convolutional Neural Networks (CNNs) to detect and alert clinicians of visible, but sometimes subtle, polyps in colonoscopy video frames \citep{hassan2020,livovsky2021,luo2021}. While these algorithms have demonstrated impressive detection rates, they require that polyps appear in the colonoscope field of view (FoV) during a procedure to be detected. Missed regions --- areas of the colon never imaged during a screening procedure --- were found to make up an estimated 10\% of the colon surface in a retrospective analysis of endoscopic video \citep{mcgill2018}, putting patients at risk of experiencing interval CRC.

To reduce the extent of missed regions, research has explored measuring observational coverage of the colon during screening procedures \citep{hong2007,hong2011,armin2016}. \cite{freedman2020} describe a  data-driven method for directly regressing a numerical visibility score for a small cluster of frames using deep learning. Some methods utilize deep learning to regress pixel-level depth \citep{mahmood2018,rau2019,cheng2021}, and this information can be incorporated into simultaneous localization and mapping (SLAM) techniques to reconstruct the colon surface \citep{chen2019a}. Holes in these reconstructions could indicate tissue areas that have gone unobserved, and these regions have been flagged in real time to alert the colonoscopist \citep{ma2019}. Other relevant applications of 3D computer vision techniques in colonoscopy include polyp size prediction \citep{abdelrahim2022}, surface topography reconstruction \citep{parot2013}, visual odometry estimation \citep{yao2021}, and enhanced lesion classification with 3D augmentation \citep{mahmood2018b}.

\subsection{Related work}

\subsubsection{Endoscopy reconstruction datasets}

Datasets for evaluating endoscopic reconstruction methods differ by intended application, model type, recording setup, and ground truth data availability. Acquiring datasets with accurate surface information in the surgical environment is generally impractical. As an alternative, both commercial \citep{stoyanov2010} and computed tomography (CT)-derived \citep{penza2018} silicone models have been explored. Surface information can be directly measured with either CT or optical scanning (OS), and the stereo sensor configuration in some laparoscopes can be used to generate ground truth depth \citep{recasens2021}. Beyond synthetic models, which suffer from relatively homogeneous optical properties, both \textit{ex-vivo} \citep{mahmood2018,edwards2020,allan2021,maier2014,ozyoruk2020} and \textit{in-vivo} \citep{ye2017} animal tissues have been used to generate data with a more realistic bidirectional scattering distribution functions (BSDF).

More recently, game engines, such as Unity (Unity Technologies) have been used to render synthetic images from 3D anatomical models, such as CT colonography volumes \citep{mahmood2018c,rau2019, ozyoruk2020, zhang2021, rau2022}. Rendered data is advantageous because error-free, pixel-level ground truth labels such as depth and surface normals are available from rendering primitives. Additionally, large quantities of data may be quickly produced. One significant drawback is the limited ability of rendering engines to simulate real-world camera optics, non-global illumination, proprietary post-acquisition processing, sensor noise, and light-tissue interaction. 

The majority of endoscopic reconstruction datasets with real images are designed for laparoscopic imaging of the abdomen and thorax. Laparoscopic datasets are unsuitable for benchmarking colonoscopic imaging because of large differences in the angular FoV (and corresponding distortion), sensor arrangement, and organ geometry. Furthermore, it is challenging to mimic the realistic motions of a colonoscope with a rigid laparoscope. Ozyoruk et al. provide a video dataset for imaging the stomach, small bowel, and colon with several different camera types with one sequence recorded using a clinical colonoscope \citep{ozyoruk2020}. Each video sequence in this dataset is paired with a ground truth camera trajectory and 3D surface model, but no pixel-level ground truth data was generated, limiting the utility of the data. Generating pixel-level ground truth information is particularly challenging for colonoscopy, due to the space-constrained imaging environment, high resolution requirements, and large range of working distances that are relevant for clinical applications.

\begin{figure}[t!]
\centering
\includegraphics[width={\linewidth}]{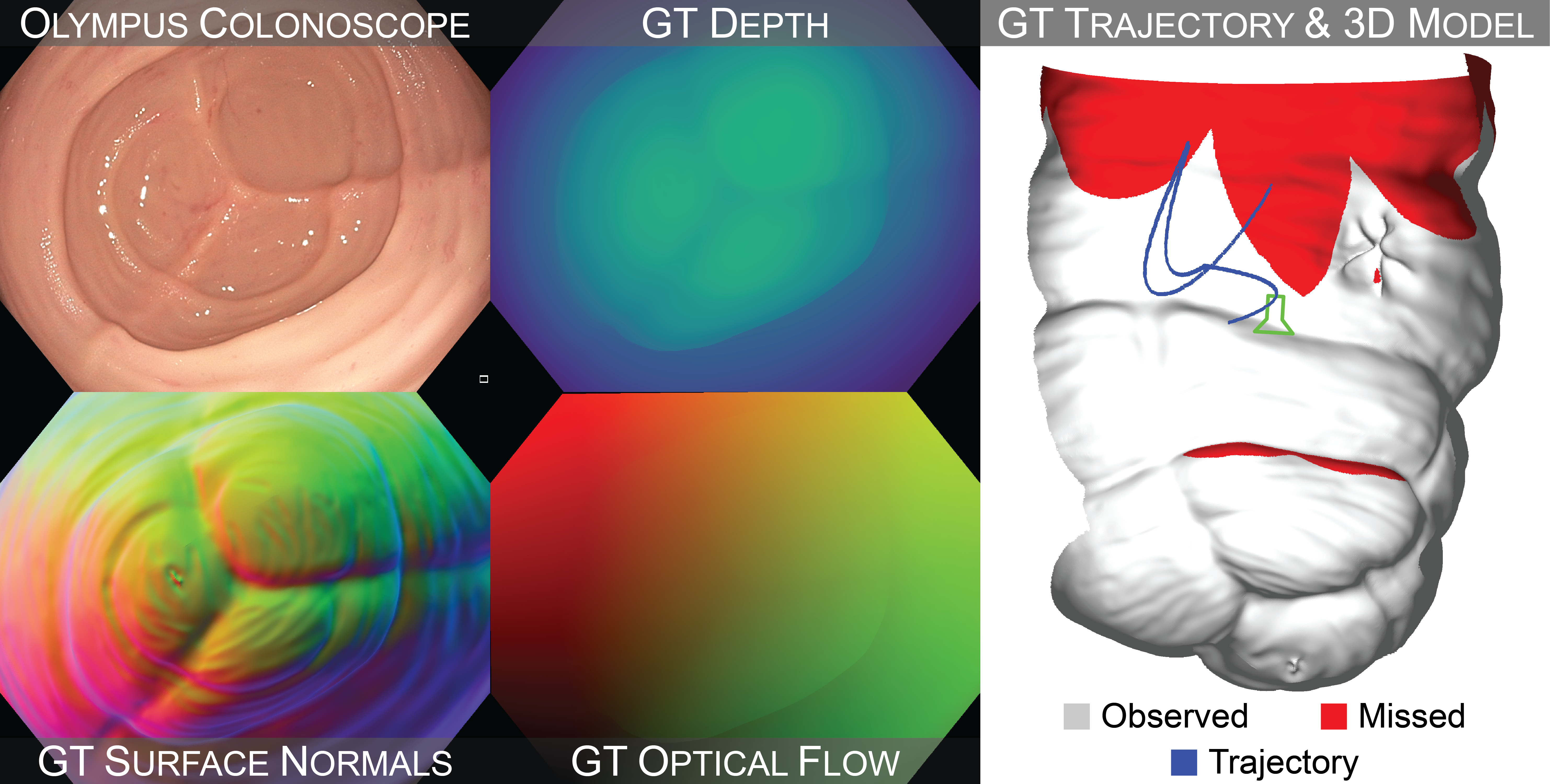}
\caption{\textbf{Sample frame from the proposed dataset.} Real colonoscope frames are paired with registered ground truth (GT) depth, surface normals, and optical flow frames (left). Each video is paired with a ground truth camera trajectory, 3D surface model, and coverage map (right).}
\label{fig:abstract}
\end{figure}

\begin{table*}[t!]
\centering
\begin{threeparttable}
\caption{\label{tab:comparisonMethods}Comparison of endoscopy reconstruction datasets}
\footnotesize
\setlength\tabcolsep{2.5pt}
\begin{tabular}{l l l l l l r c c c c c l}
\toprule
\multicolumn{3}{c}{\textbf{Camera}} & & & & \multicolumn{6}{c}{\textbf{Ground Truth}} & \\
\cmidrule(lr){1-3} \cmidrule(lr){7-12}
\textbf{Type} & \textbf{FoV} & \textbf{Res.} & \textbf{Tissue Type} & \textbf{3D GT} & \textbf{Registration} & \textbf{Frames} & \textbf{Pose} & \textbf{Depth} & \textbf{Normals} & \textbf{Flow} & \textbf{3D Model} & \textbf{Dataset} \\
\midrule
Stereo da Vinci     & Narrow  & SD      & Ex-vivo Porcine & CT      & Manual        & 16      & \checkmark  &             &            &            & \checkmark & \cite{edwards2020}         \\
Stereo CMOS         & Narrow  & SD      & Phantom         & CT      & Calib Plate   & 120     &             & \checkmark  &            &            & \checkmark & \cite{penza2018}           \\
Stereo da Vinci     & Narrow  & SD      & Phantom         & CT      & Fiducials     & 5,816   &             &             &            &            & \checkmark & \cite{stoyanov2010};       \\
                    &         &         &                 &         &               &         &             &             &            &            &            & \cite{pratt2010}           \\
Stereo da Vinci     & Narrow  & SD      & In-vivo Animal  & None    & None          & 20,000  &             &             &            &            &            & \cite{ye2017}              \\
Stereo da Vinci     & Narrow  & SD      & In-vivo Animal  & None    & None          & 92,672  &             & \checkmark  &            &            & \checkmark & \cite{recasens2021}        \\
Stereo da Vinci     & Narrow  & SD      & Ex-vivo Porcine & SI      & None          & 40      &             & \checkmark  &            &            &            & \cite{allan2021}\tnote{*}  \\
Stereo da Vinci     & Narrow  & SD      & Ex-vivo Porcine & CT      & Fiducials     & 131     &             &             &            &            & \checkmark & \cite{maier2014}           \\
Stereo Rendered     & Narrow  & SD      & Virtual         & CT      & None          & 6,320   & \checkmark  &             &            &            &            & \cite{zhang2021}           \\
Mono USB            & Narrow  & SD      & Phantom         & None    & None          & 23,935  & \checkmark  &             &            &            &            & \cite{fulton2020}          \\
Mono Rendered       & Narrow  & SD      & Virtual         & CT      & None          & 16,016  &             & \checkmark  &            &            &            & \cite{rau2019}             \\
Mono Rendered       & Narrow  & SD      & Virtual         & CT      & None          & 18,000  & \checkmark  & \checkmark  &            &            &            & \cite{rau2022}\tnote{*}    \\
Mono USB            & Wide    & SD/HD   & Ex-vivo Porcine & OS      & None          & 39,406  & \checkmark  &             &            &            & \checkmark & \cite{ozyoruk2020}         \\
Mono Pill Cam       & Wide    & SD      & Ex-vivo Porcine & None    & None          & 3,294   & \checkmark  &             &            &            &            & \cite{ozyoruk2020}         \\
Mono Colonoscope    & Wide    & HD      & Phantom         & CT      & None          & 12,250  & \checkmark  &             &            &            & \checkmark & \cite{ozyoruk2020}         \\
Mono Rendered       & Narrow  & SD      & Virtual         & CT      & None          & 21,887  & \checkmark  & \checkmark  &            &            &            & \cite{ozyoruk2020}         \\
\textit{Mono Colonoscope} & \textit{Wide} & \textit{HD} & \textit{Phantom} & \textit{Sculpted} & \textit{Optimized} & \textit{10,015} & \checkmark & \checkmark & \checkmark & \checkmark & \checkmark & \textit{Proposed} \\
\bottomrule
\end{tabular}
\begin{tablenotes}
\centering{\item[]Ground Truth (GT), Computed Tomography (CT), Structured Illumination (SI), Optical Scan (OS)}
\item[*] Only available to challenge participants
\end{tablenotes}
\end{threeparttable}
\end{table*}

\subsubsection{Registering real and virtual endoscopy images}
Registering endoscopic frames with ground truth surface models enables derivative ground truth data and metrics to be extracted. For example, ground truth depth frames may be assigned to real endoscopy images provided the endoscope pose relative to the surface model is known. However, registration using conventional, feature-based methods is challenging due to a lack of robust corner points and variable specular reflections common in endoscopic images. To circumvent this challenge, segmented fiducials have been used to register optical images to ground truth CT volumes \citep{rau2019,stoyanov2010}. While this method is robust, one drawback is the presence of unrealistic fiducials throughout the image FoV. \cite{edwards2020} opt to remove the fiducials and instead rely on the manual alignment of a virtual camera with a ground truth CT volume. The manual nature of this method works well for producing small quantities of data, but it is a barrier to registering large quantities of data, and its accuracy is limited by inter-operator variability. \cite{penza2018} use a calibration target to calibrate the coordinate systems of a fixed camera and laser scanner, allowing for simultaneous recording of endoscopy frames and 3D surface information in laparoscopic imaging environments that are not size-constrained.

A summary of existing and the proposed endoscopic 3D datasets and acquisition methods is reported in Table \ref{tab:comparisonMethods}.

\subsubsection{2D-3D registration}

2D-3D registration enables the registration of a 2D image with a 3D spatial volume, and it is frequently used for aligning 3D preoperative CT volumes with 2D intraoperative X-ray images. Most methods rely on optimizing the similarity between a target 2D image and a simulated 2D radiograph of the 3D volume acquired at the estimated pose \citep{markelj2012}. Common similarity measures are gradient- \citep{livyatan2003}, intensity- \citep{birkfellner2003}, and feature-based \citep{groher2007} metrics. More recently, 2D-3D registration methods have evolved to include learning-based algorithms to address challenges in cross-modality registration \citep{oulbacha2020} and feature extraction \citep{grupp2020}.

\subsection{Contributions}

While the body of computer vision research in colonoscopy is extensive, evaluating and benchmarking methods remains a challenge due to a lack of ground truth annotated data. In this work, we present a High Definition (HD) \underline{C}olonoscopy \underline{3}D \underline{V}ideo \underline{D}ataset (C3VD) for quantitatively evaluating computer vision methods. To the best of our knowledge, this is the first video dataset with 3D ground truth that is recorded entirely with a clinical colonoscope. In this work, we contribute:

\begin{enumerate}
\item{a 2D-3D video registration algorithm for aligning real 2D optical colonoscopy video sequences with ground truth 3D models. GAN-estimated depth frames are compared with rendered predicted views along a measured camera trajectory for minimizing an edge-based loss.}
\item{a technique for generating high-fidelity silicone phantom models with varying textures and colors to facilitate domain randomization. }
\item{a ground truth dataset with pixel-level registration to a known 3D model for quantitatively evaluating computer vision techniques in colonoscopy.  This dataset contains 10,015 HD video frames of realistic colon phantom models obtained with a clinical colonoscope. Each frame is paired with ground truth depth, surface normals, occlusion, optical flow, and a six degree-of-freedom camera pose. Each video is paired with a ground truth surface model and coverage map (Figure \ref{fig:abstract}).}
\end{enumerate}

The dataset, 3D colon model, phantom molds, and 2D-3D registration algorithm are all made publicly available at \url{durr.jhu.edu/C3VD}.

\section{Methods} \label{sec:methods}

We propose a method for generating video sequences through a clinical colonoscope with paired, pixel-level ground truth. We first describe a protocol for producing high-fidelity phantom models (Section \ref{sec:phantoms}) and recording video sequences with ground truth trajectory (Section \ref{sec:dataAquisition}). We then introduce a novel technique for registering the acquired video and trajectory sequences with a ground truth 3D surface model (Section \ref{sec:registration}).

\subsection{Phantom model production} \label{sec:phantoms}

A complete 3D colon model - from sigmoid colon to cecum - was digitally sculpted in Zbrush (Pixologic) by a board-certified anaplastologist (JRG) using reference anatomical imagery from colonoscopic procedures. This method introduces higher frequency detail in the digital models as compared to models derived from resolution-limited CT colonography. We split the sculpted model into five segments: the sigmoid colon, descending colon, transcending colon, ascending colon, and cecum. Three-part molds were generated for each segment. Two parts comprised two halves of the outer shell and one insert part formed the colon lumen and mucosal surface. All molds were 3D printed with an Objet 260 Connex 3 with 16 micrometer resolution. Casts of each mold were created with silicone (Dragon Skin\textsuperscript{TM}, Smooth-On, Inc.). Silicone pigments (Silc Pig\textsuperscript{TM}, Smooth-On, Inc.) were used to vary the color and texture. Silicone was manually applied in 5-12 layers with varying degrees of opacity to emulate patient-specific tissue features and vasculature patterns at varying optical depths. A silicone lubricant (015594011516, BioFilm, Inc.) was applied to the surface of the models at recording time to simulate the highly specular appearance of the mucosa.

\begin{figure}[!t]
\centerline{\includegraphics[width=\linewidth]{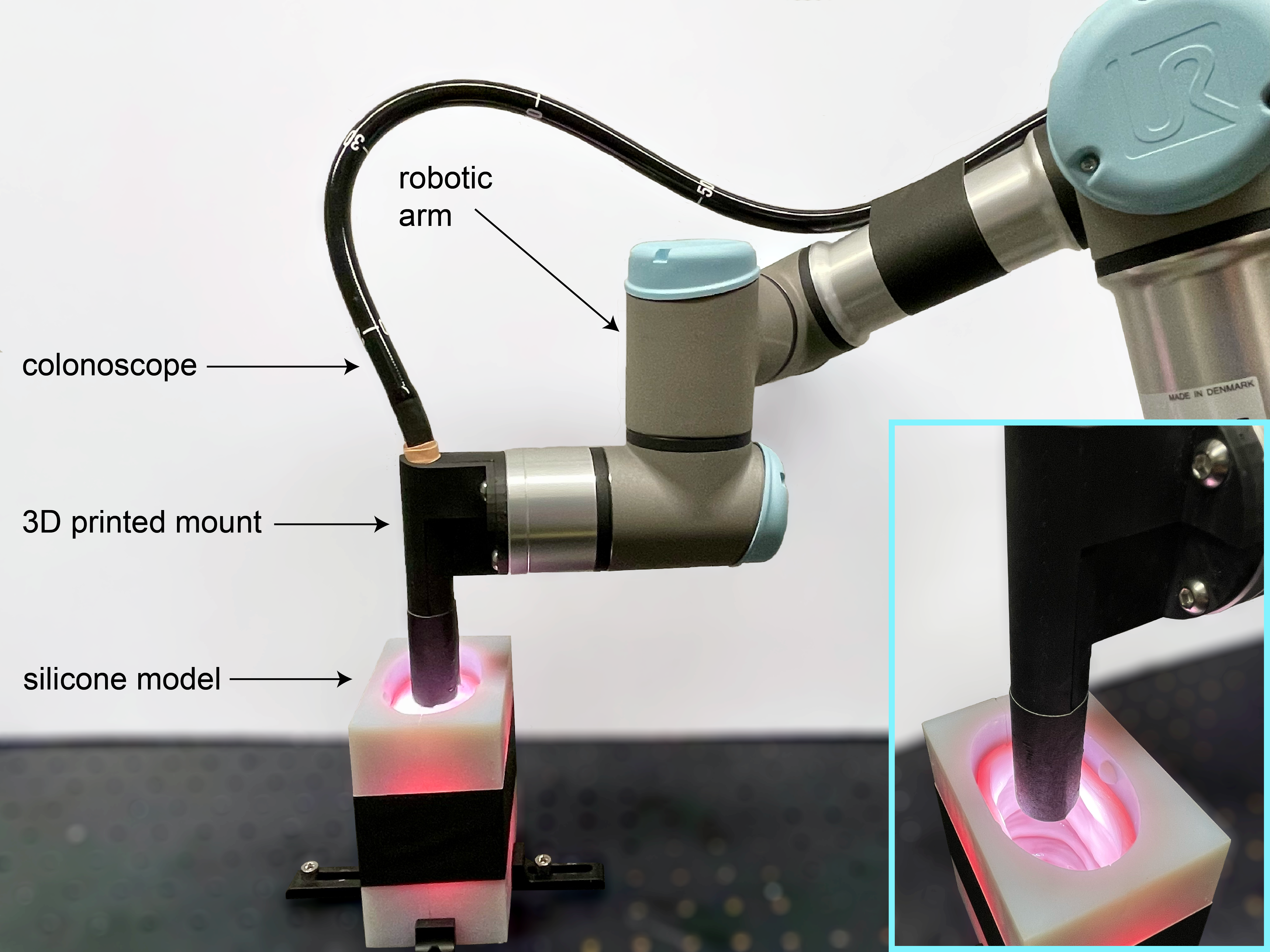}}
\caption{\textbf{Data acquisition setup.} A commercial colonoscope is rigidly affixed to a robotic arm using a 3D printed mount. The colonoscope is navigated through the silicone colon with known 3D shape while video and pose are simultaneously recorded.}
\label{fig:setup}
\end{figure}

\subsection{Data acquisition} \label{sec:dataAquisition}

The setup used for data acquisition is shown in Figure \ref{fig:setup}. An Olympus CF-HQ190L video colonoscope, CV-190 video processor, and CLV-190 light source were used to record video sequences of the phantom models. The models were placed inside the molds with inserts removed to keep them static and free of deformation during video recording. The tip of the colonoscope was rigidly mounted to a UR-3 (Universal Robotics) robotic arm. For each video segment, 4-8 sequential poses were manually programmed to mimic a typical colonoscopy trajectory. The robotic arm traversed the colon with interpolation between these poses with 10 micrometer repeatability. A pose log was recorded from the arm at a sampling rate of 63 Hz. Colonoscopy videos were recorded in an uncompressed HD format using an Orion HD (Matrox) frame grabber connected to the SDI output of the Olympus video processor.

\subsection{Registration pipeline overview} \label{sec:registration}

\begin{figure*}[t!]
\centerline{\includegraphics[width=\textwidth]{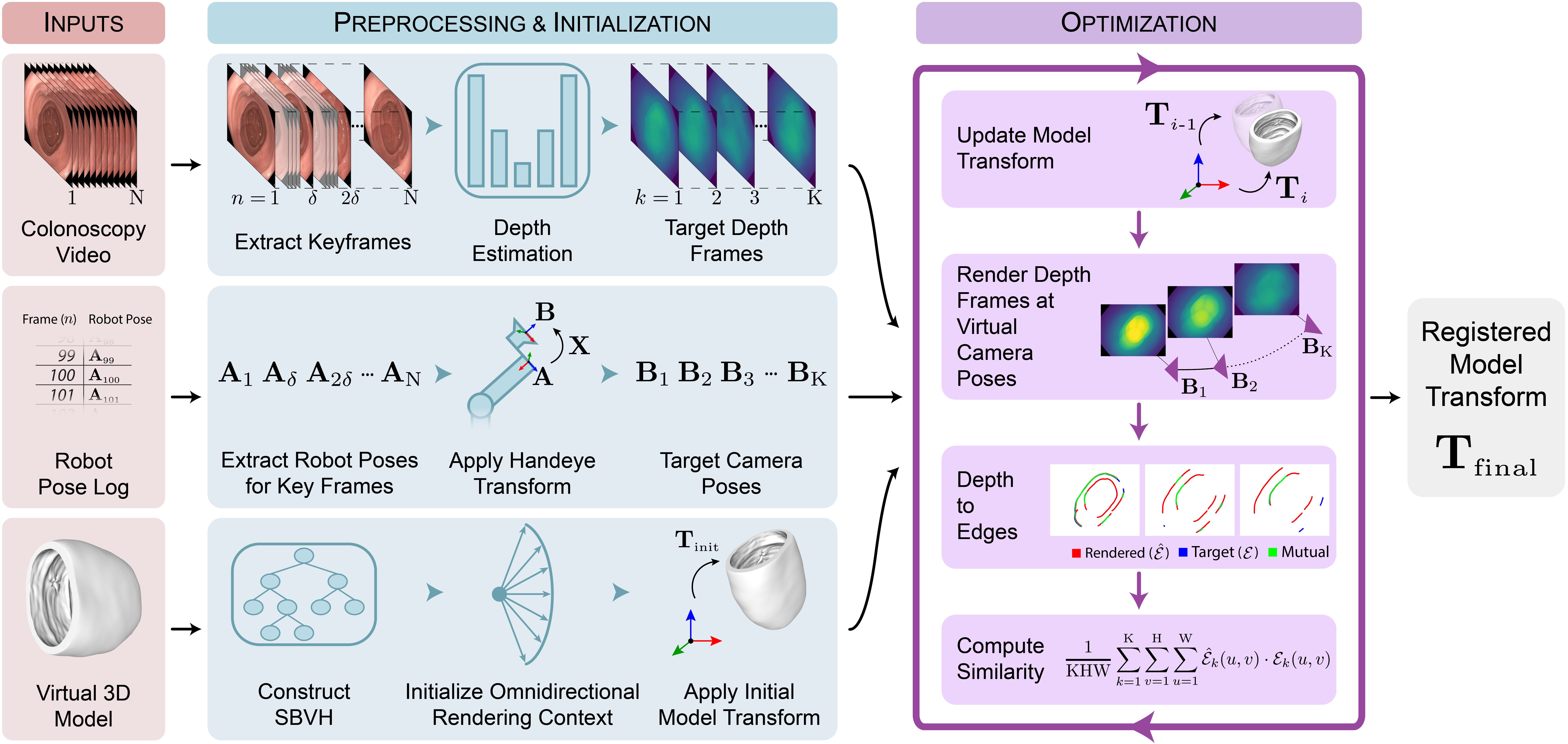}}
\caption{\textbf{2D-3D video registration method.} Temporally synchronized video and pose measurements are sampled to produce a set of keyframes for registration with the ground truth 3D model. Keyframes are individually transformed to a depth domain by a generative model to produce target depth images. A virtual omnidirectional camera is moved to each keyframe pose to render depth images of the 3D model.  Rendered and target depth frames are compared while the model transform is updated with an evolutionary optimizer until convergence.}
\label{summary}
\end{figure*}

Pixel-level ground truth for each video frame was generated by moving a virtual camera along the recorded trajectory and rendering ground truth frames of the 3D model. While the colonoscope trajectory for the virtual camera was known, the location of the phantom model relative to this trajectory was unknown. Assuming the phantom model was stationary for the duration of a video sequence, the phantom pose can be expressed as a single rigid body transform ($\mathbf{T}_{\mathrm{final}}$), consisting of a rotational component ($\mathbf{\theta}$) and translational component ($\mathbf{t}$). We parameterize this transformation using three Euler angles and a translation vector ($[\mathbf{\theta}_\alpha\;\mathbf{\theta}_\beta\;\mathbf{\theta}_\gamma\;\mathbf{t}_x\;\mathbf{t}_y\;\mathbf{t}_z]$). To estimate this unknown transform, we utilize a 2D-3D registration approach to align geometric features shared between the 2D video frames and the virtual 3D model of the phantom. The registration pipeline samples the parameter space for a model transform prediction ($\mathbf{T}_i$), evaluates the feature alignment between target depth frames and renderings at the current model transform, and updates the model transform prediction using an evolutionary optimizer. This registration method is outlined in Figure \ref{summary} and detailed in the following subsections.

\subsubsection{Data preprocessing} \label{sec:preprocess}

Before beginning the registration optimization, each video sequence, pose log, and ground truth 3D model were preprocessed. For a video sequence consisting of N frames, keyframes were sampled at an interval of $\delta$, resulting in K total keyframes. A depth frame was then estimated for each keyframe using a Generative Adversarial Network trained to estimate depth (Section \ref{sec:depthgan}).

For each keyframe, a pose $\mathbf{A}_i$ describing the position of the robotic arm end-effector relative to the base was sampled from the pose log. Synchronization between the video sequence and pose log was achieved by solving for the relative temporal offset that resulted in the maximum correlation between the optical flow magnitude and the pose displacement of the camera. A handeye calibration was performed to characterize the transformation,  $\mathbf{X}$,  between robotic arm pose, $\mathbf{A}_{i}$, and colonoscope camera pose ($\mathbf{B}_i$). This calibration allows the relationship

\begin{equation}
    \mathbf{A}_{a}^{-1}\mathbf{A}_{b}\mathbf{X}=\mathbf{X}\mathbf{B}_{a}^{-1}\mathbf{B}_{b},
\end{equation}

\noindent where $\mathbf{A}_{a}$/$\mathbf{B}_{a}$ and $\mathbf{A}_{b}$/$\mathbf{B}_{b}$ are pairs of robot and camera poses captured for calibration. Once calibrated, robotic arm poses were transformed to camera poses by solving a rearranged version of the handeye relationship 

\begin{equation}
    \mathbf{B}_i = \mathbf{B}_{cal}\mathbf{X}^{-1}\mathbf{A}_{cal}^{-1}\mathbf{A}_{i}\mathbf{X},
\end{equation}

\noindent where $\mathbf{A}_{cal}$ and $\mathbf{B}_{cal}$ are a pair of poses retained from the calibration.

Finally, the ground truth triangulated mesh was converted to a Split Bounding Volume Hierarchy (SBVH), and a rendering context was created for rendering depth frames from the 3D ground truth model (Section \ref{sec:rendering}). The initial model transform ($\mathbf{T}_{\mathrm{initial}}$) for each video sequence was manually aligned. We used a custom graphical user interface that overlayed the first keyframe with frames rendered at camera pose $\mathbf{B}_1$ as the model transform was manually perturbed.

\subsubsection{Target depth estimation} \label{sec:depthgan}
Predicting pixel-level depth for each keyframe was formulated as an image-to-image translation task. A conditional generative adversarial network (cGAN) was trained with synthetically rendered input-output image pairs, and inference was performed on real images. Previous studies demonstrating strong domain generalizability when trained on synthetic data and applied to real data motivated using a cGAN network architecture \citep{chen2018,rau2019}. 1,000 pairs of synthetic colonoscopy images with paired depth were rendered using the virtual 3D models and 3 unique BSDFs for domain randomization. The descending colon model and a fourth BSDF were omitted from the training data and saved for validation experiments. To enable training at HD-resolution, multi-scale discriminator models and a multi-layer feature matching loss were employed in addition to the traditional GAN loss \citep{wang2018}. Keyframes were fed to the trained generator to produce target depth frames for alignment.

\subsubsection{Rendering depth frames} \label{sec:rendering}

\begin{figure}
\centerline{\includegraphics[width={0.8\linewidth}]{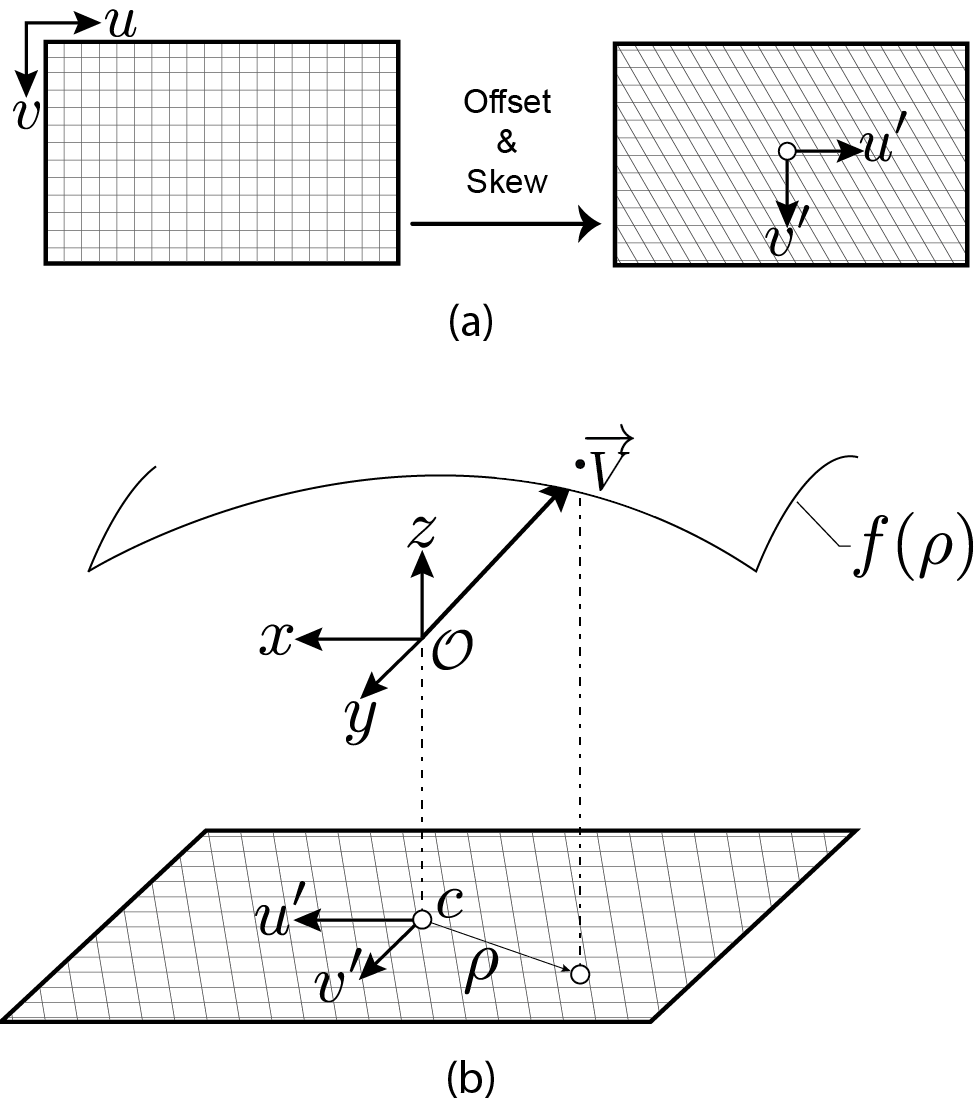}}
\caption{\textbf{Omnidirectional camera model.} (a) Pixel coordinates are skewed by an affine transform to model lens misalignment and offset to the optical center. (b) Each pixel is assigned a ray direction defined by a parametric surface that is a function of the pixel's radial distance from the optical center.}
\label{fig:omnicam}
\end{figure}

Depth frames at each camera pose were rendered using a virtual camera in the 3D colon model, and these frames were compared to the target depth frames to evaluate the accuracy of the current model transform prediction. A wide FoV ($\approx$170\textdegree) is a key characteristic of clinical colonoscopes that maximizes the visibility of the peripheral colon surface. However, this fisheye effect has not been simulated in synthetic endoscopy datasets \citep{ozyoruk2020,rau2022}. To model the entire FoV of a commercial colonoscope, we adopted a spherical camera intrinsic model \citep{scaramuzza2006} as shown in Figure \ref{fig:omnicam}.

For each pixel $[u\:v]$, there exists a ray direction $\overrightarrow{V}\in\mathbb{N}^{3}$ emanating from the camera origin $\mathcal{O}\in\mathbb{R}^{3}$ into world space. Pixel coordinates are first shifted with respect to the optical center of the camera $c\in\mathbb{R}^{2\times1}$, and then skewed with an affine transformation $\mathrm{A}\in\mathbb{R}^{2\times2}$ to account for lens misalignment, producing distorted pixel coordinates $[u'\:v']\in\mathbb{R}^{2}$:

\begin{equation}
    \begin{bmatrix} u' \\ v' \end{bmatrix} =
    \mathrm{A}^{-1}
    \begin{bmatrix} u-c_x \\ v-c_y \end{bmatrix},\;
    \mathrm{A} = \begin{bmatrix} e & f \\ g & 1 \end{bmatrix}.
\end{equation}

Distorted pixel coordinates and corresponding ray directions are related by a parametric equation

\begin{equation}
    \begin{bmatrix}\overrightarrow{V}_x \\ \overrightarrow{V}_y \\ \overrightarrow{V}_z\end{bmatrix} =
    \begin{bmatrix} u' \\ v' \\ f(\rho) \end{bmatrix},
\end{equation}

\noindent where $\rho= \sqrt{u'^2 + v'^2}$, $f(\rho)=\alpha_0 + \alpha_2\rho^2 + \alpha_3\rho^3 + \alpha_4\rho^4$, and $\alpha_0 ... \alpha_4$ are coefficient parameters solved for during calibration. $\overrightarrow{V}$ is transformed by the rotational matrix for the current camera pose and the resulting ray direction is cast from the ray origin.

For each optimization iteration, the model transform is first updated to the current sample ($\mathbf{T}_{i-1} \rightarrow \mathbf{T}_{i}$), and depth frames ($\mathcal{D}_{\mathbf{T}}$) are rendered at each virtual camera pose ($\mathbf{B}_1 \cdots \mathbf{B}_{\mathrm{K}}$). For efficient computation, we implement a custom raycasting engine using the Nvidia\textregistered{} OptiX\textsuperscript{TM} raytracing API. This enables direct access to the GPU RT cores \citep{parker2010}. The model SBVH is used to initialize an acceleration structure, and the structure is updated with each new model transform.

\subsubsection{Edge loss \& optimization}

Target and rendered depth frames are compared to evaluate the accuracy of the current prediction. To minimize the effect of scale inconsistency that is common in depth predicted by deep learning, we compare geometric contours from depth discontinuities in the target and model depth frames. Contours are extracted using Canny edge extraction and binarized. To provide a continuous and smooth loss function when comparing target and model depth frames, we blurred these edges with a normalized Gaussian kernel, denoting this operation as $\hat{\mathcal{E}} = E(\mathcal{D})$. The registration optimization then aims to maximize the overlap of these blurred edge frames by computing a similarity metric between frame pairs,

\begin{equation} \label{eqn:objective}
sim(\hat{\mathcal{E}},\mathcal{E}) = \frac{1}{\textrm{KHW}}\sum_{k=1}^{\textrm{K}} \sum_{v=1}^{\textrm{H}} \sum_{u=1}^{\textrm{W}} \hat{\mathcal{E}}_k(u,v) \cdot \mathcal{E}_k(u,v),
\end{equation}

\noindent where $\hat{\mathcal{E}}$ is the set of rendered edge frames, $\mathcal{E}$ is the set of target edge frames, and $\mathrm{K}$ is the number of frame pairs. The output of this similarity function is a value between 0.0 and 1.0 that is maximal when the target and rendered edges are equivalent and perfectly aligned. To reframe the function as a minimization problem, the similarity value is subtracted from 1.0, yielding the full objective function

\begin{equation} \label{eqn:fullObjective}
\mathbf{T}_{\mathrm{final}} = \argmin{\mathbf{T}}(\:1.0 - sim(E(\mathcal{D}_{\mathbf{T}}),\mathcal{E})\:),
\end{equation}

Evaluation samples for the model transform are iteratively refined by an evolutionary optimizer called Covariance Matrix Adaptation Evolution Strategy (CMA-ES) \citep{hansen2003}.

\subsection{Simulated screening colonoscopies} \label{sec:simulation}

In addition to the video segments with pixel-level ground truth that can be computed with robot-arm fine pose information, we recorded four simulated screening colonoscopy video sequences with coarse six degree-of-freedom pose and 3D surface models. The phantom model segments were adhered together with silicone adhesive (Sil-Poxy\textsuperscript{TM}, Smooth-On, Inc.) and mounted in a laser cut foam scaffold. An electromagnetic field generator (Aurora, Northern Digital Inc.) was positioned above the model, and a six degree-of-freedom electromagnetic sensors (EM, 610016, Northern Digital Inc.) was rigidly affixed to the distal tip of the scope for recording pose information at 40Hz. Beginning at the cecum, the scope was withdrawn by a trained gastroenterologist (VSA) while video and pose information was recorded. The EM poses were synchronized and transformed to camera poses using the same process described in Section \ref{sec:preprocess}. If the sensor failed to track for a portion of the trajectory, temporally neighboring poses were linearly interpolated.

\begin{figure}
\centerline{\includegraphics[width={0.87\linewidth}]{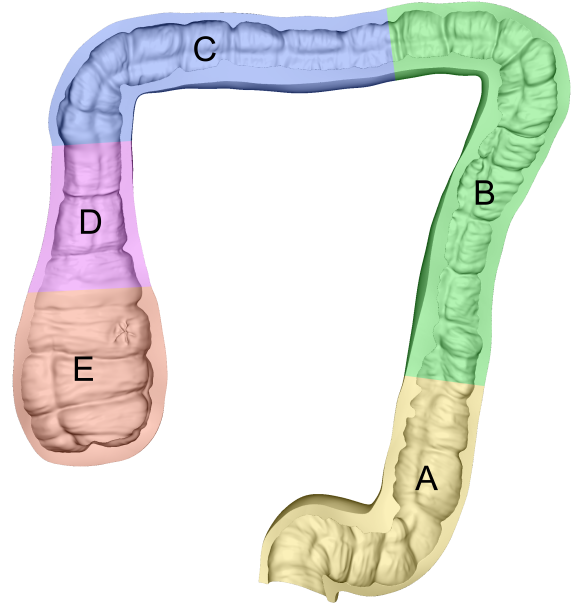}}
\caption{\textbf{Cross section of the ground truth 3D colon model.} Sculpting the 3D model using reference anatomical images introduces higher-resolution detail than CT colonography volumes.}
\label{fig:model}
\end{figure}

\begin{table}[t]
\centering
\begin{threeparttable}
\caption{\label{tab:segments}Ground truth 3D colon model attributes}
\footnotesize
\begin{tabularx}{\linewidth}{llll}
\toprule
\textbf{Segment}  & \textbf{Length (mm)}    & \textbf{Lesion Type}    & \textbf{Major axis (mm)}    \\
\midrule
A                 & 192                     & Hyperplastic            & 4.2                         \\
                  &                         & Traditional adenomatous & 6.6                         \\
\midrule
B                 & 339                     & Sessile serrated        & 6.2                         \\
\midrule
C                 & 241                     & Sessile serrated        & 10.0                        \\
\midrule
D                 & 93                      & -                       & -                           \\
\midrule
E                 & 97                      & -                       & -                           \\
\bottomrule
\end{tabularx}
\end{threeparttable}
\end{table}

\section{Experiments and results} \label{sec:results}

\subsection{Implementation}
A cross-section of the complete sculpted colon model is shown in Figure \ref{fig:model}. Model segment lengths as well as lesion types and major axis diameters are reported in Table \ref{tab:segments}. The 3D printed molds for this model were used to cast four complete sets of phantoms. Each phantom set was cast with unique material properties. 

Camera intrinsic parameters for the CF-HQ190L video colonoscope were measured using 30 images of a 10x15 checkerboard with 10x10 millimeter squares and the Matlab 2022a Camera Calibration Toolbox (Mathworks). The calibration resulted in a 0.47 pixel mean reprojection error. The final results of this calibration are reported in Table \ref{tab:intrinsics}. The camera extrinsic poses estimated from the calibration were used with the corresponding robot poses to compute the handeye transform $\mathbf{X}$. The solution for $\mathbf{X}$ was found using the optimization method proposed by \cite{park1994}.

\begin{table}[t]
\centering
\begin{threeparttable}
\caption{\label{tab:intrinsics}Omnidirectional camera intrinsics for Olympus CF-HQ190L video colonoscope}
\footnotesize
\begin{tabularx}{\linewidth}{XXX}
\toprule
Size            & H$\times$W &  $1080\times1350$     \\
\midrule
Optical Center  & $c_x$      &  $679.54$             \\
                & $c_y$      &  $543.98$             \\
\midrule
Polynomial      & $\alpha_0$ &  $769.24$             \\
                & $\alpha_2$ &  $-8.13\times10^{-4}$ \\
                & $\alpha_3$ &  $-6.26\times10^{-7}$ \\
                & $\alpha_4$ &  $-1.20\times10^{-9}$ \\
\midrule
Stretch         & $e$        &  $0.9999$             \\
                & $f$        &  $2.88\times10^{-3}$  \\
                & $g$        &  $-2.96\times10^{-3}$ \\
\bottomrule
\end{tabularx}
\end{threeparttable}
\end{table}

The depth estimation network architecture was implemented in Pytorch. The initial learning rate of 0.0002 was held constant for the first 50 epochs, then linearly decayed for the final 50 epochs. Input frames were zero-padded to a size of $1088 \times 1376$ and fed to the network with a batch size of 1. The training computation was split across 4 Nvidia\textregistered{}  RTX A5000 Graphics Processing Units, taking 31 hours to complete. Real colonoscopy video frames were then input into the trained generator to produce target depth frames. Because the target depth frames remain constant during registration optimization, the target edge features are extracted once before beginning iterative optimization.

The complete registration pipeline was implemented in C++ and CUDA and computed using an Nvidia\textregistered{}  RTX 2080 TI Graphics Processing Unit. Sample evaluations were rendered, edges extracted, and a loss computed at an average rate of 4.0 milliseconds per keyframe.

\begin{figure}
\centerline{\includegraphics[width=\linewidth]{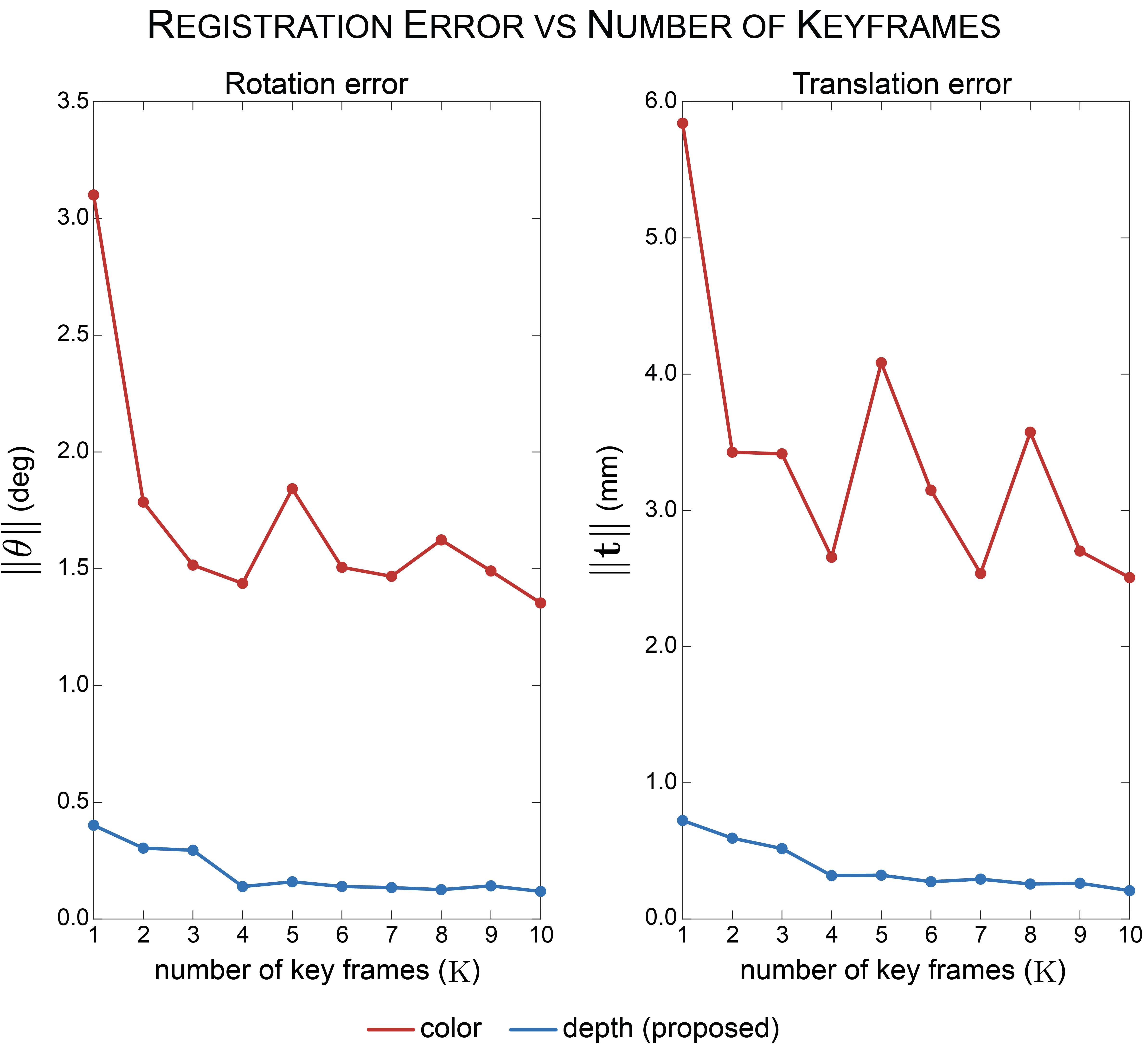}}
\caption{\textbf{Average registration error for 10 synthetic video sequences with increasing number of keyframes K.} The registration error decreases with increasing number of keyframes. Registering frames in the depth domain (blue line) results in improved registration accuracy compared to registering frames in the color domain.}
\label{fig:deltaPlot}
\end{figure}

\begin{table}[t]
\centering
\begin{threeparttable}
\caption{\label{tab:losses}Effect of loss functions on registration accuracy}
\footnotesize
\begin{tabularx}{\linewidth}{XXXX}
\toprule
\textbf{Input} & \textbf{Loss} & \textbf{$\lVert \mathbf{\theta}_{\mathrm{error}} \rVert$ (deg)} & \textbf{$\lVert \mathbf{t}_{\mathrm{error}} \rVert$ (mm)}\\
\midrule
Depth         & $L_1$             & 8.008            & 14.941            \\
              & $L_2$             & 8.729            & 16.109            \\
              & NCC               & 2.904            & 6.014             \\
              & GC                & 0.558            & 0.843             \\
\midrule
Edge          & $L_1$             & 0.466            & 0.654             \\
              & $L_2$             & 0.417            & 0.750             \\
              & DICE              & 0.374            & 0.756             \\
              & NCC               & 0.450            & 0.681             \\
              & \textbf{Proposed} & \textbf{0.159}   & \textbf{0.321}    \\
\bottomrule
\end{tabularx}
\end{threeparttable}
\end{table}

CMA-ES iterations were computed with a population size of 100 samples. The parameter space was bounded to $\theta=\pm0.1$ radians and $\mathbf{t}=\pm7.5$ millimeters with respect to ($\mathbf{T}_{\mathrm{initial}}$). The bounds were linearly scaled to a uniform space during parameter sampling ($\sigma=0.1$) to account for the difference in scale between rotation and translation parameters.

\subsection{Evaluation with synthetic data}
To quantitatively validate the registration error of the proposed algorithm, 10 synthetic video sequences, each composed of 200 frames, were rendered with a 3D model and a BSDF not included in the dataset used to train the depth estimation network. In each experiment, the model initial transform was first manually aligned, then optimized using the registration pipeline. To quantify the registration accuracy, we compute the error $\mathbf{T}_{\mathrm{error}}$ as

\begin{equation} \label{eqn:error}
\mathbf{T}_{\mathrm{error}} = (\mathbf{T}_{\mathrm{gt}})^{-1} \: \mathbf{T}_{\mathrm{final}}\:,
\end{equation}

\noindent where $\mathbf{T}_{\mathrm{gt}}$ is the ground truth model transform and $\mathbf{T}_{\mathrm{final}}$ is the final registered transform, both in homogeneous form. The homogeneous transform $\mathbf{T}_{\mathrm{error}}$ is broken into rotation and translation components, converting the rotation component to Euler form ($\mathbf{\theta}$), then independently evaluating $\lVert \mathbf{\theta} \rVert$ and $\lVert \mathbf{t} \rVert$.

\begin{figure*}[t!]
\centerline{\includegraphics[width=0.97\textwidth]{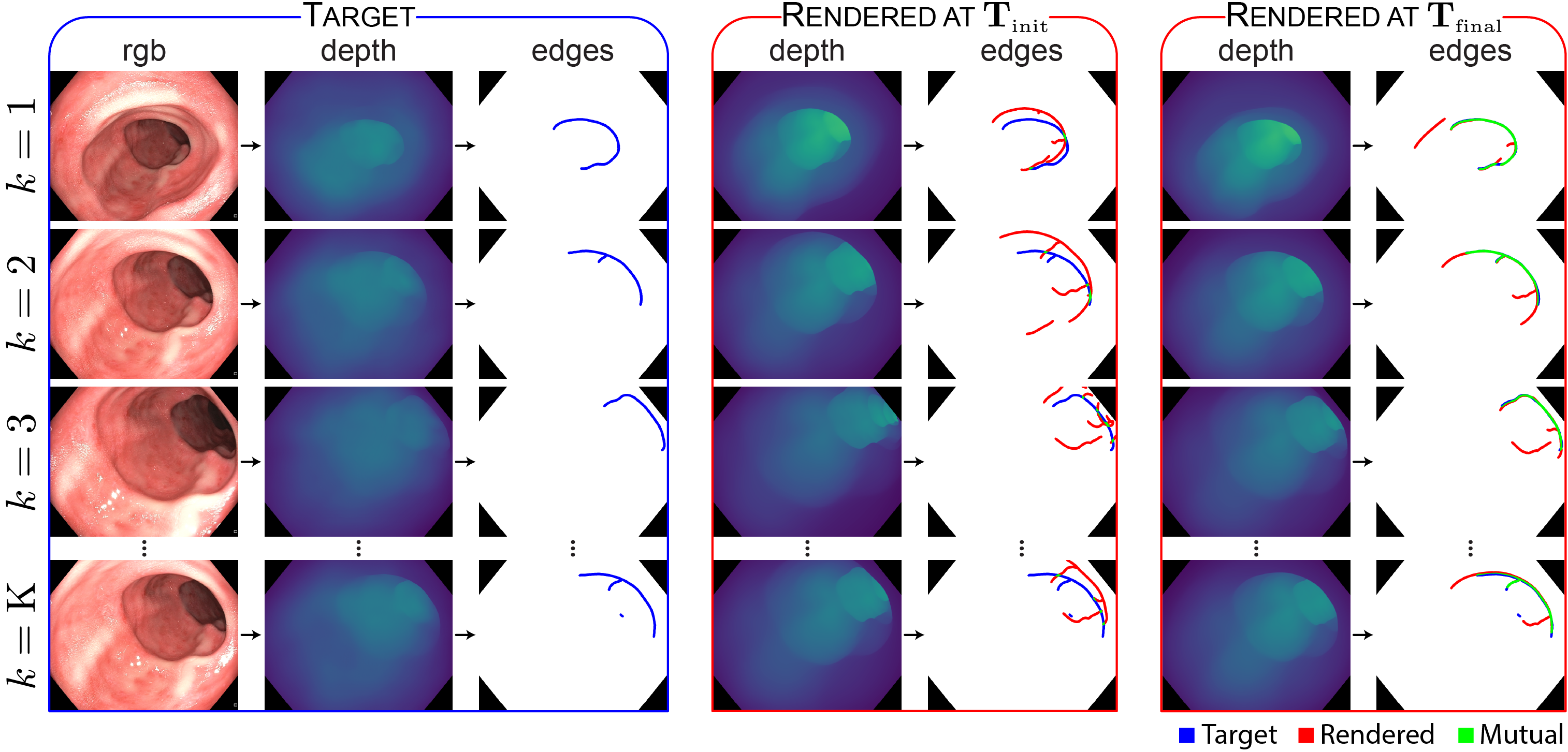}}
\caption{\textbf{Qualitative registration results.} RGB video frames are transformed to the depth domain by a generative model, and target edge features are extracted. Edge features from rendered depth frames are compared with the target edge features to compute an alignment loss, and the predicted model transform is iteratively updated until alignment is achieved. Blurred edge features are binarized in this figure for visibility. See Supplementary Video I for a visual depiction of the improvement in registration as the optimization progresses.}
\label{fig:edges}
\end{figure*}

\begin{figure*}[t!]
\centerline{\includegraphics[width=0.97\textwidth]{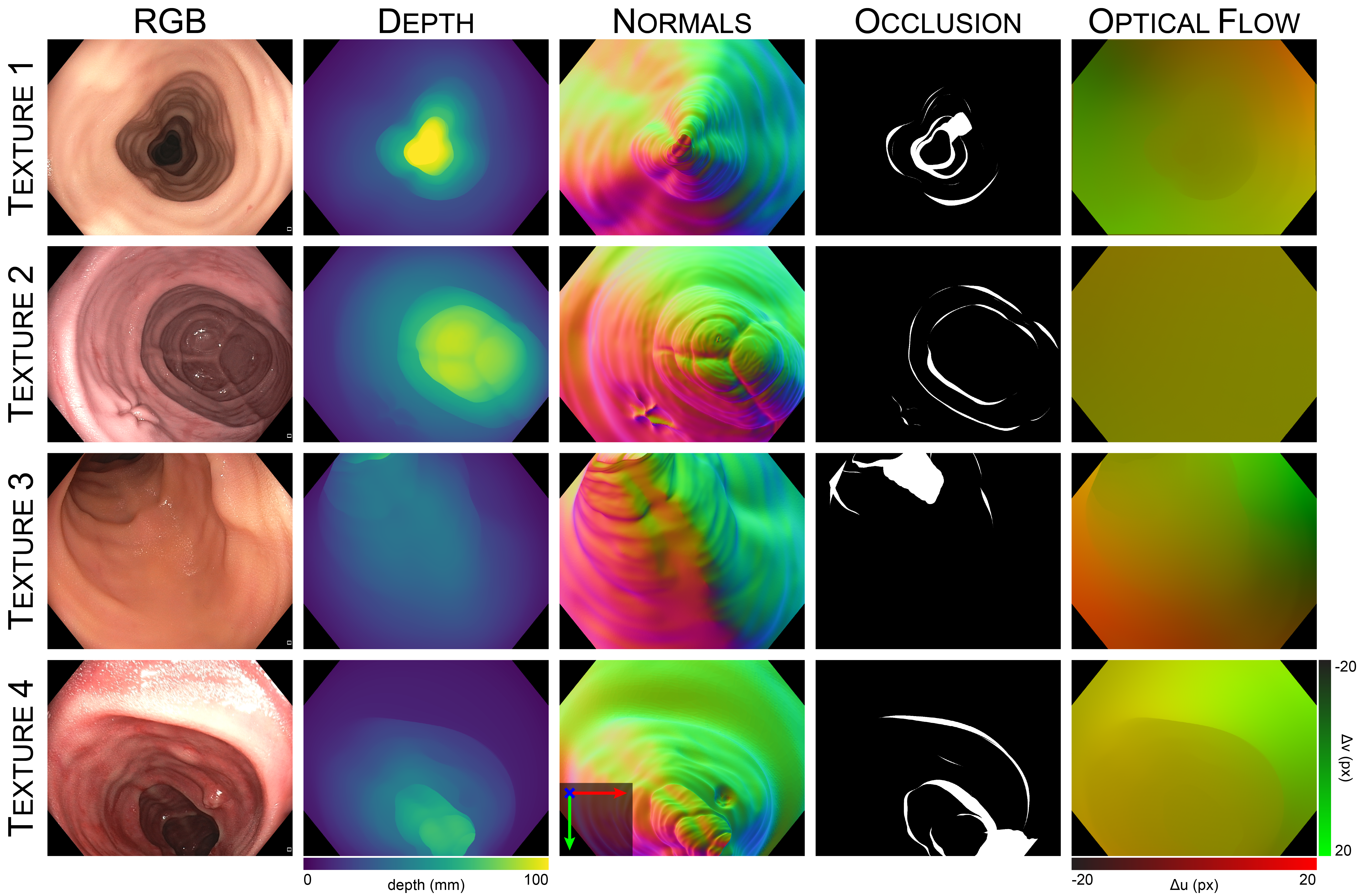}}
\caption{\textbf{Sample frames from the C3VD dataset.} Frames are recorded using a clinical colonoscope and silicone phantom models constructed with four unique textures. Each real video frame is paired with ground truth depth, surface normals, occlusion, and optical flow. A sample video sequence is shown in Supplementary Video II.}
\label{fig:sampleFrames}
\end{figure*}

Using these synthetic sequences and validation metrics, we first investigated the effect of number of keyframes $\mathrm{K}$ on registration accuracy. Optimizations were run for all 10 video sequences with $\mathrm{K}$ ranging from 1 to 10. Results are presented in Figure \ref{fig:deltaPlot}. We find that the rate of improvement in registration accuracy significantly decreases at $\mathrm{K}$ equal to 5, achieving a 0.321 millimeter and 0.159 degree accuracy. This resulted in an improvement of the translational accuracy by 55.6\% and rotational accuracy by 60.4\% compared to using only a single frame. With this in mind, $\mathrm{K}$ is set equal to 5 for the remaining experiments to balance registration accuracy with time-efficiency of computation.

To evaluate the effect of using blurred edge features in the loss function, we experimented with optimizing the alignment of depth frames prior to edge extraction. Depth-based registrations were optimized using $L_1$, $L_2$, Normalized Cross Correlation (NCC), and Gradient Correlation (GC) loss functions \citep{desilva2016}. To validate our loss function selection for edges, we also experimented with $L_1$, $L_2$, DICE (using binarized edges), and NCC losses with edge frames. The results for each input type and loss function are reported in Table \ref{tab:losses}.

To assess the benefit of registering in the depth domain, we registered synthetic video frames without depth transformation to predicted views rendered with a Lambertian BSDF. Edges were extracted from the color synthetic frames and rendered predicted views and a loss was computed using the proposed function. This method resulted in an average registration accuracy that was an order of magnitude larger than registration in the depth domain (Figure \ref{fig:deltaPlot}).

Lastly, we examined how trajectory complexity affects registration accuracy. Synthetic colonoscopy sequences were rendered using three trajectory types with increasing complexity: simple (linear translation only), moderate (helical translation only), and complex (helical translation with pitch and yaw rotation). 10 synthetic sequences per trajectory type were rendered and registered to the ground truth 3D model. This experiment resulted in 0.117-degree, 0.063-degree, and 0.070-degree rotational error for simple, moderate, and complex trajectories, respectively. The translational errors were 0.150-millimeter, 0.070-millimeter, and 0.089-millimeter for simple, moderate, and complex trajectories, respectively.

\subsection{Evaluation with real video sequences}
Once the registration pipeline was validated using synthetic data with available ground truth, the algorithm was then applied to real recorded videos and poses. While the ground truth model transform for real data is unknown, the quality of each registration may be evaluated by qualitatively comparing the edge alignment quality across each video. Sample results for a recorded sequence are reported in Figure \ref{fig:edges}. The results demonstrate that the registration pipeline aligns sample frames with the target frames throughout the entire video.

\begin{figure}[t!]
\centerline{\includegraphics[width=\linewidth]{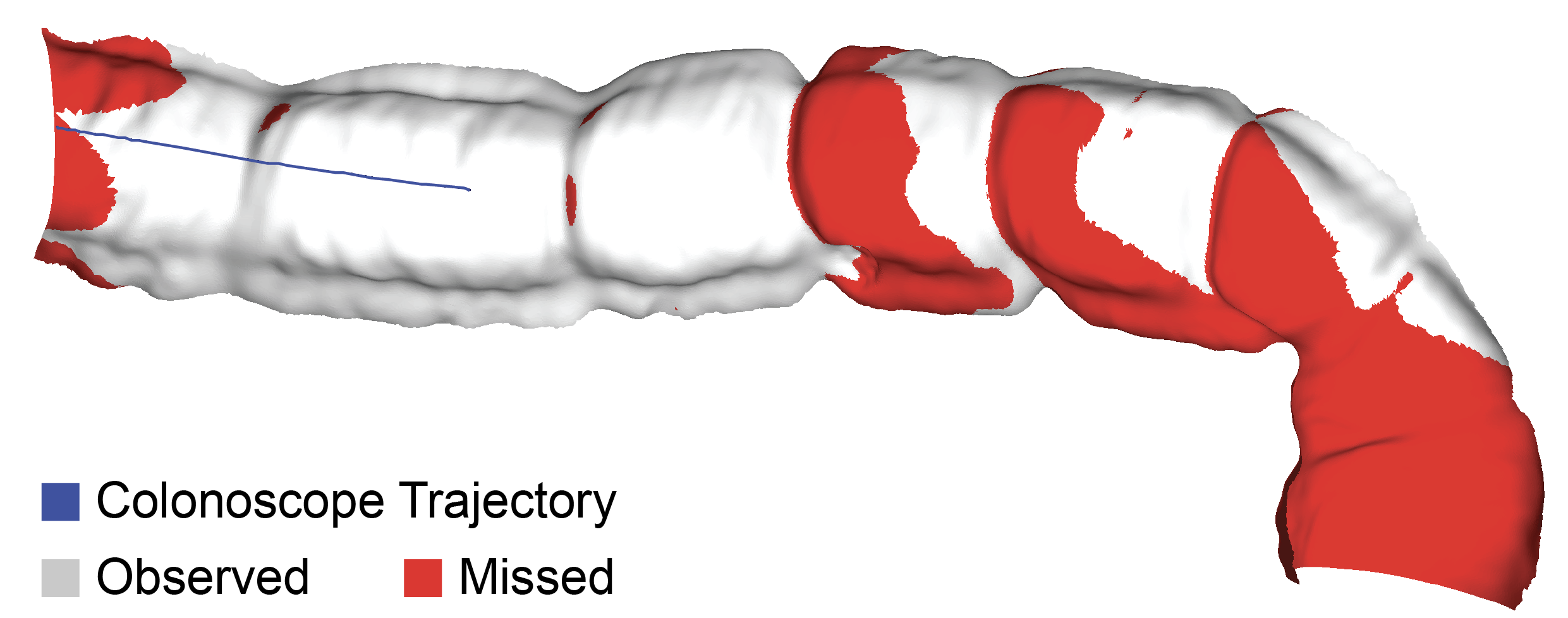}}
\caption{\textbf{Sample coverage map for a recorded video sequence.} Unobserved surface regions due to occlusion are marked as red.}
\label{fig:coverage}
\end{figure}

\section{Dataset and distribution format}\label{sec:format}
22 video sequences consisting of 10,015 frames were registered for inclusion in the dataset. The illumination conditions in each sequence were varied by adjusting the illumination mode (automatic versus manual) and illumination power settings on the clinical light source. The sequences also include a combination of “down-the-barrel”, “en-face”, and partially occluded views. A listing of the dataset video sequences is included in Table \ref{tab:dataset}.
Once registered, the virtual camera was moved to the camera pose corresponding with each video frame, and ground truth depth, surface normals, occlusion, and optical flow were rendered for each frame. Sample dataset frames are displayed in Figure \ref{fig:sampleFrames}. For each video sequence, a coverage map was generated by accumulating the surface faces observed in each frame throughout the video, then marking those faces which went unobserved. A sample coverage map for a video sequence from the dataset is shown in Figure \ref{fig:coverage}.
For every frame in the video dataset, we provide a corresponding:
\vspace{-0.15cm}
\begin{itemize}
 \item Depth frame: Depth along the camera frame's $z$-axis, clamped from 0-100 millimeters. Values are linearly scaled and encoded as a 16-bit grayscale image.
\vspace{-0.15cm}
\item Surface normal frame: Reported with respect to the camera coordinate system. X/Y/Z components are stored in separate R/G/B color channels. Components are linearly scaled from $\pm$1 to 0-65535. Values are encoded as a 16-bit color image.
\vspace{-0.15cm}
\item Optical flow frame: Computed flowing from the current frame to the previous frame, meaning the first frame in the sequence has no value: $I_{i-1}  = I_i(u+\Delta u,v+\Delta v)$. Values are saved in a color image, where the R-channel contains X-direction motion (left $\rightarrow$ right, -20 to 20 pixels), and the G-channel contains Y-direction motion (up $\rightarrow$ down, -20 to 20 pixels). Values are linearly scaled from 0 to 65535 and encoded as a 16-bit color image.
\vspace{-0.15cm}
\item Occlusion frame: Encoded as an 8-bit binary image. Pixels occluding other mesh faces within 100 millimeters of the camera origin are assigned a value of 255, and all other pixels are assigned a value of 0.
\vspace{-0.15cm}
\item Camera pose: Saved in homogeneous form.
\end{itemize}
\vspace{-0.15cm}
For each video sequence, we also provide:
\vspace{-0.15cm}
\begin{itemize}
\item 3D model: Ground truth triangulated mesh, stored as a Wavefront OBJ file.
\vspace{-0.15cm}
\item Coverage map: Binary texture indicating which mesh faces were observed by the camera during the video sequence (1=observed, 2=unobserved).
\end{itemize}
\vspace{-0.15cm}
Finally, we include four video sequences of simulated screening procedure, consisting of 20,058 frames, and ground truth camera poses in homogeneous form.

\begin{table}[t!]
\centering
\begin{threeparttable}
\caption{\label{tab:dataset}C3VD dataset video sequences}
\footnotesize
\begin{tabularx}{\linewidth}{XXXX}
\toprule
\textbf{Segment}  & \textbf{Texture}    & \textbf{Video}  & \textbf{Frames} \\
\midrule
A                 &  1                  & a               & 700             \\
                        &  2                  & a               & 514             \\
                        &  3                  & a               & 613             \\
                        &  3                  & b               & 536             \\
\midrule
B                 &  4                  & a               & 148             \\
\midrule
C                 &  1                  & a               & 61              \\
                        &  1                  & b               & 700             \\
                        &  2                  & a               & 194             \\
                        &  2                  & b               & 103             \\
                        &  2                  & c               & 235             \\
                        &  3                  & a               & 250             \\
                        &  3                  & b               & 214             \\
                        &  4                  & a               & 382             \\
                        &  4                  & b               & 597             \\
\midrule
E                 &  1                  & a               & 276             \\
                        &  1                  & b               & 765             \\
                        &  2                  & a               & 370             \\
                        &  2                  & b               & 1,142           \\
                        &  2                  & c               & 595             \\
                        &  3                  & a               & 730             \\
                        &  3                  & b               & 465             \\
                        &  4                  & a               & 425             \\
\midrule
A-E\tnote{*}      &  1                  & a               & 5,458           \\
                        &  2                  & a               & 5,100           \\
                        &  3                  & a               & 4,726           \\
                        &  4                  & a               & 4,774           \\
\bottomrule
\end{tabularx}
\begin{tablenotes}
\centering{\item[*] Simulated screening colonoscopies}
\end{tablenotes}
\end{threeparttable}
\end{table}

\section{Discussion and conclusion} \label{sec:discuss}

This work presents a 2D-3D registration method for acquiring and registering phantom colonoscopy video data with ground truth surface models. Unlike traditional 2D-3D registration techniques that register single views, this method operates on video sequences and measured pose. Our results demonstrate that leveraging this temporal information improves the registration accuracy compared to registering a single frame (Figure \ref{fig:deltaPlot}). To circumvent registration errors caused by specular reflections and surface textures, we transformed keyframes to a depth domain for similarity evaluation. The transformed depth frames show scale inconsistency (Figure \ref{fig:edges}), as is common in GAN-predicted depth. We found the impact of this inconsistency to be reduced by aligning edge features extracted from the depth frames, further improving the registration accuracy (Table \ref{tab:losses}). Additionally, our loss function outperforms other metrics such as NCC and DICE, while also being computationally cheaper. 

We also found that the registration accuracy for moderate and complex trajectories outperformed simple trajectories. One possible explanation is an increase in the number of unique edge features used for alignment and their increased translation in the imaging plane in moderate and complex trajectories. Prior to registration optimization, the input sequence is sampled to generate a set of keyframes, and this number of keyframes is constant (K=5) for all trajectories. While simple trajectories included only linear translation primarily in the colon major axis direction, moderate and complex trajectories also included lateral translation as well as pitch and yaw variation. The camera poses in complex trajectories are more diverse than simple trajectories, which results in a larger diversity of edge features for alignment. Additionally, simple linear translation of the camera parallel to the major colon axis results in much smaller displacements in image features compared to camera translation perpendicular to the colon axis.

This registration method was used to generate C3VD, the first colonoscopy reconstruction dataset that includes real colonoscopy videos labeled with registered ground truth. Unlike the EndoSLAM dataset \citep{ozyoruk2020}, C3VD is recorded entirely with an HD clinical colonoscope and includes depth, surface normal, occlusion, and optical flow frame labels. While EndoSLAM uses real porcine tissue mounted on non-tubular scaffolds, we opt for tubular phantom models to simulate a geometrically realistic lumen. By recording with a real colonoscope, C3VD overcomes the limited ability of renderers to simulate non-global illumination, light scattering, and non-linear post-processing \citep{rau2022}. Compared to other datasets which omit large portions of the angular FoV, C3VD is the first dataset to model the entire colonoscope FoV through an omnidirectional camera model.

C3VD can be used to validate the sub-components of SLAM reconstruction algorithms, including estimating pixel-level depth, surface normals, optical flow, and occluded regions. The ground truth surface models and coverage maps can be used to evaluate entire SLAM reconstruction methods and techniques for identifying missed regions. Additionally, the 3D model assets are open-sourced and offer a higher-resolution alternative to CT colonography volumes for rendering synthetic training data. Mold files and fabrication protocols are also provided so that other researchers may produce and modify phantom colon models. The simulated screening colonoscopy videos may be used to test algorithms on full sequences with realistic scope motion.

The proposed methods have several important limitations. Precise pose measurements are required to generate the registered video sequences, and these trajectories are recorded by mounting the colonoscope to a robotic arm. While this method enables several additional forms of ground truth information, the scope range and types of motion are limited. Additionally, the phantom models must remain static during video acquisition, whereas colonic tissue is flexible, dynamic, and in frequent contact with the scope during imaging. Lastly, small errors in the handeye calibration accumulate, causing a gradual drift in the camera trajectory. Future work could improve upon these limitations with adjustments to the data acquisition protocol and registration algorithm. A more-accurate radiofrequency positional sensor could circumvent trajectory limitations imposed by the robotic arm. Handeye calibration errors could possibly be reduced by co-optimizing both the model transformation and the handeye transformation. This method could also be paired with dynamic CT, similar to \cite{stoyanov2010}, to produce video data with deformable colon models.

The C3VD dataset presented here also has important limitations in its diversity and scope. Future work could expand this dataset to encompass a larger variety of endoscopic imaging and tissue conditions. For quality metric applications, such as fractional coverage, additional trajectories and colon shapes with a variety of missed regions should be developed. Among the challenges to applying 3D reconstruction algorithms to colonoscopy are loose stool and debris that often occlude the field of view. To simulate these artifacts, artificial stool could be applied to the silicone models to span varying levels of bowel preparation quality. The auxiliary water jet and suction systems of the colonoscope could be used to simulate bowel cleaning and underwater imaging in the videos. Similarly, the air and water nozzles may be used to remove debris from the objective lens. Future datasets may also benefit from including alternative endoscope systems and modalities, such as narrow band imaging (NBI) and chromoendoscopy. The use of distal attachments such as EndoCuff \citep{rex2018} and the effect on observational coverage could also be explored. Finally, we also envision generating ground truth data for evaluating structured illumination \cite{parot2013} and hyperspectral methods \citep{yoon2021}. Validation data for these methods could include phantoms with tuned optical scattering and absorption properties to realistically simulate light-tissue interactions \citep{ayers2008,chen2019b,sweer2019}.

\section*{Acknowledgments}
This work was supported in part with funding from the National Institutes of Health Trailblazer Award (R21 EB024700) and the National Science Foundation Graduate Research Fellowship Program (DGE-1746891). This work was also supported in parts with funding and products provided by Olympus Corporation of the Americas. Although the agreement states a Sponsored Research Agreement, Olympus is funding, but not sponsoring this research. The authors thank Dr. Jeffrey Siewerdsen (Biomedical Engineering) for research resources provided via the I-STAR Lab and Carnegie Center for Surgical Innovation at Johns Hopkins University. The authors also thank Ingo Wald (Nvidia\textregistered{}) for technical assistance with OptiX\textsuperscript{TM}.

\bibliographystyle{model2-names.bst}\biboptions{authoryear}

\end{document}